\documentclass[10pt,twocolumn,letterpaper]{article}

\usepackage{cvpr}
\usepackage{times}
\usepackage{epsfig}
\usepackage{graphicx}
\usepackage{amsmath}
\usepackage{amssymb}

\usepackage[breaklinks=true,bookmarks=false]{hyperref}

\cvprfinalcopy 

\def\cvprPaperID{****} 

\begin{document}

\title{Research on AI Composition Recognition Based on Music Rules}

\author{Yang Deng\\
NetEase Cloud Music\\
{\tt\small dynamo@cug.edu.cn}
\and
Ziyao Xu\\
Malong Technologies\\
{\tt\small ziyxu@malong.com}
\and
Li Zhou\\
China University of Geosciences\\
{\tt\small zhouli@cug.edu.cn}
\and
HuaPing Liu\\
NetEase Cloud Music\\
{\tt\small liuhuaping@corp.netease.com}
\and
AnQi Huang\\
NetEase Cloud Music\\
{\tt\small huanganqi01@corp.netease.com}}

\maketitle


	\begin{abstract}
	The development of artificial intelligent composition has resulted in the increasing popularity of machine-generated pieces, with frequent copyright disputes consequently emerging. There is an insufficient amount of research on the judgement of artificial and machine-generated works; the creation of a method to identify and distinguish these works is of particular importance. Starting from the essence of the music, the article constructs a music-rule-identifying algorithm through extracting modes, which will identify the stability of the mode of machine-generated music, to judge whether it is artificial intelligent. The evaluation datasets used are provided by the Conference on Sound and Music Technology(CSMT). Experimental results demonstrate the algorithm to have a successful distinguishing ability between datasets with different source distributions. The algorithm will also provide some technological reference to the benign development of the music copyright and artificial intelligent music.
	
	\textbf{Keywords:} 	AI composition, melody arrangement, machine music creation, mode recognition
\end{abstract}

\begin{figure*}[htb]
	\begin{center}
		\includegraphics[width=0.6\textwidth]{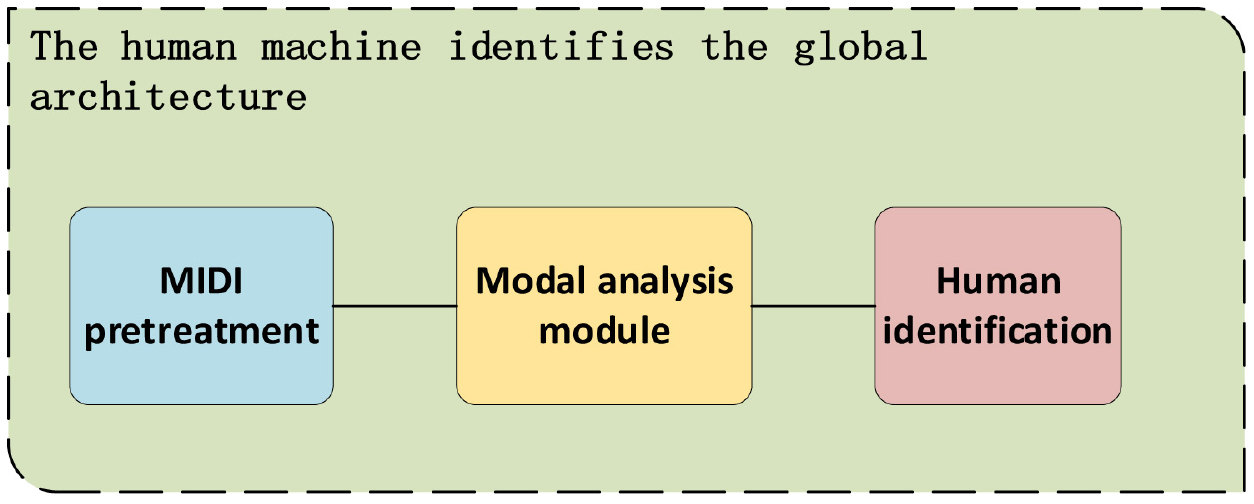}
	\end{center}
	\caption{The overall technical pattern of our approach}
	\label{fig:1}
\end{figure*}
\begin{figure*}[htb]
	\centering
	\includegraphics[width=0.8\textwidth]{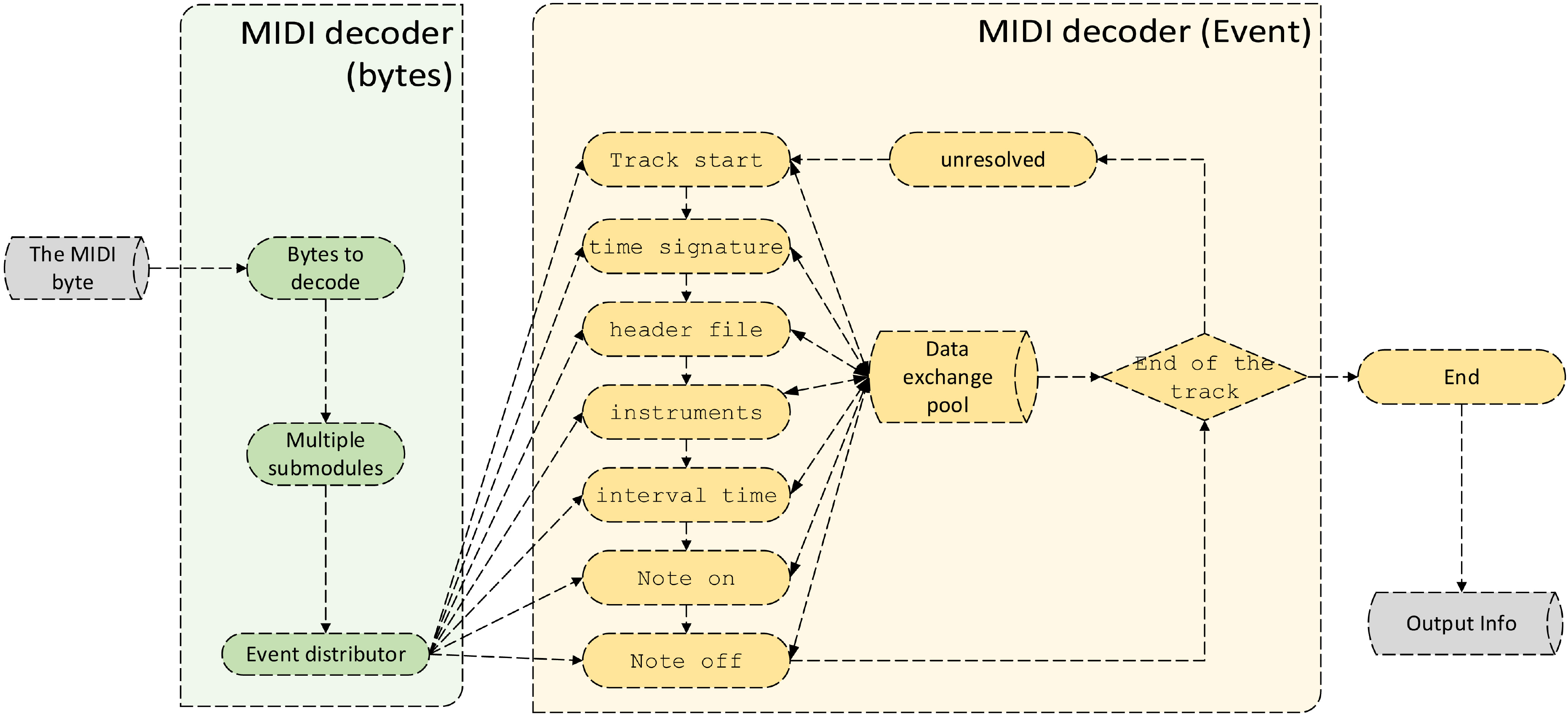}
	\caption{MIDI File preprocessing (1)}
	\label{fig:2}
\end{figure*}
\begin{figure*}[htb]
	\centering
	\includegraphics[width=0.8\textwidth]{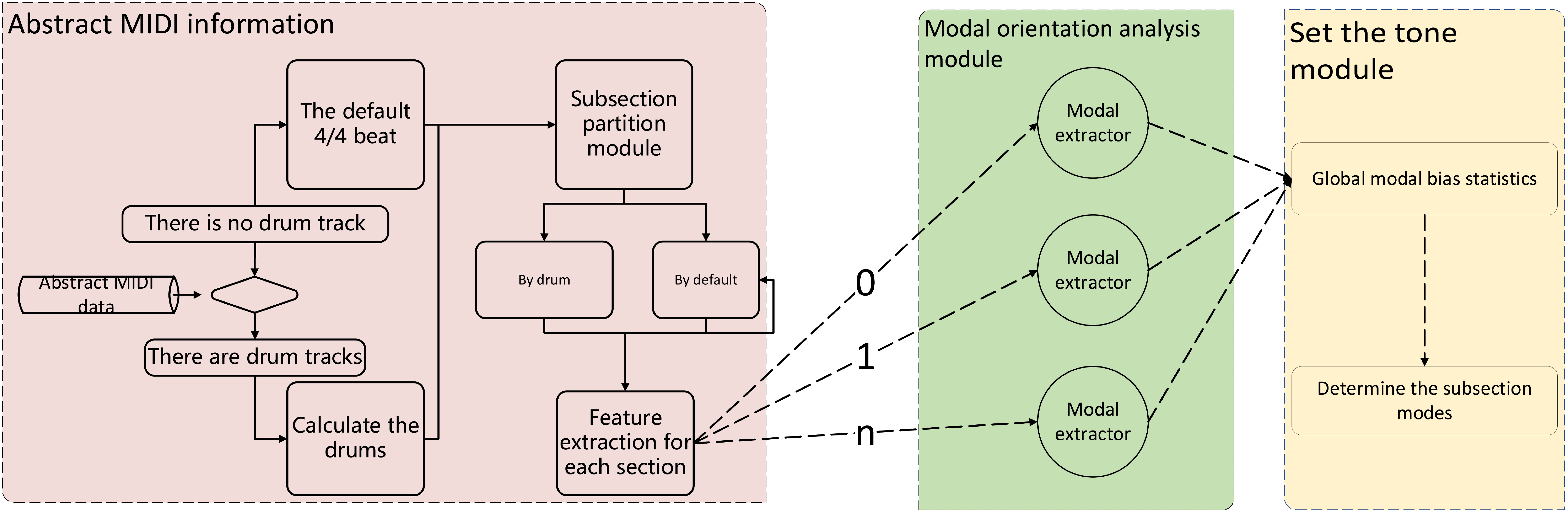}
	\caption{MIDI File preprocessing (2)}
	\label{fig:3}
\end{figure*}

\begin{figure*}[htb]
	\centering
	\includegraphics[width=0.6\textwidth]{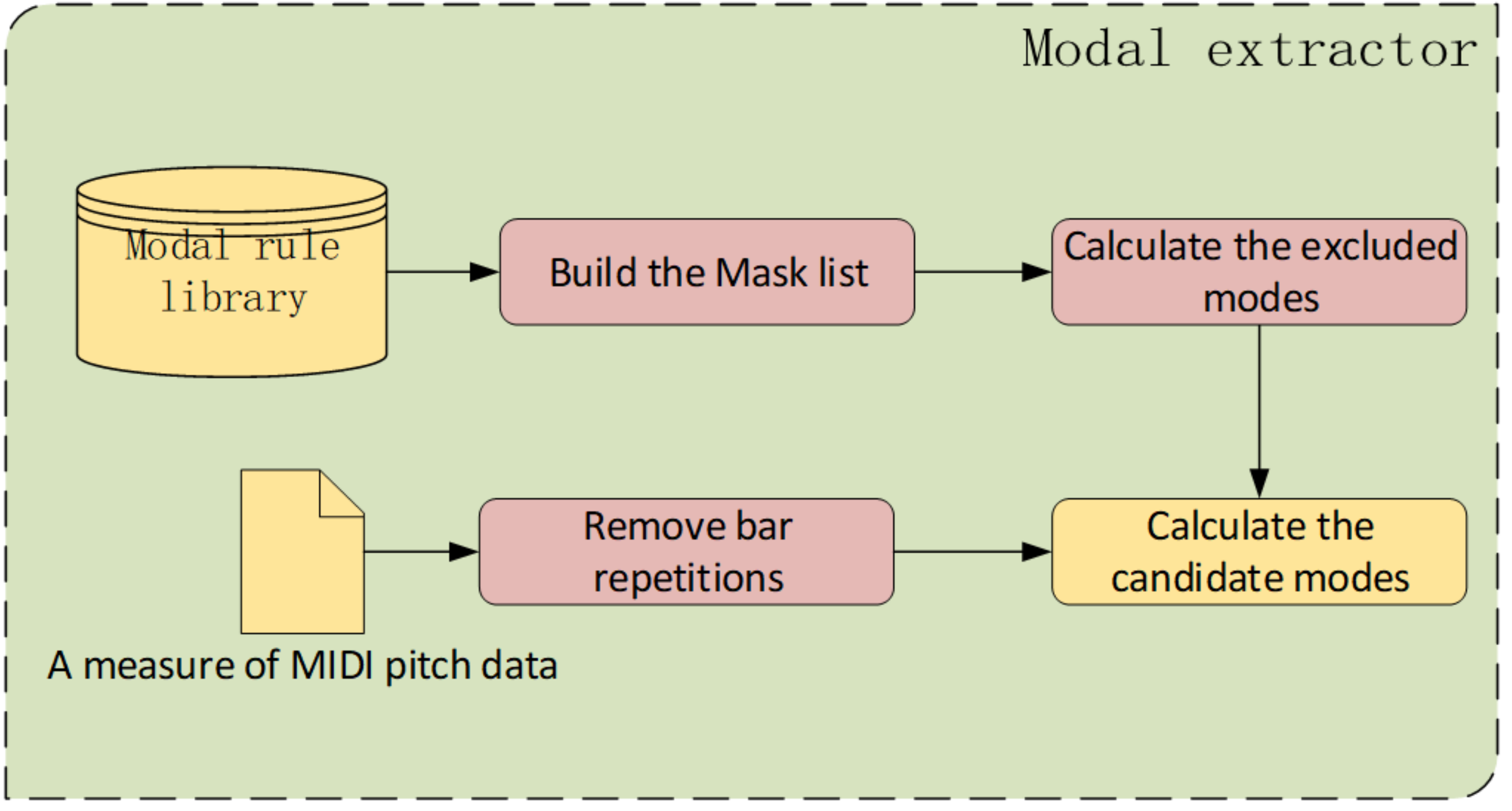}
	\caption{Modal extractor}
	\label{fig:4}
\end{figure*}
	
\section{Introduction}

With the gradual rise of artificial intelligent composition, more and more artificial intelligent composition technology has been introduced for application in the sphere of business. This technology can potentially trigger a series of disputes over copyright issues. For the purpose of managing these potential challenges to intellectual property, it is crucial to design an algorithm that can distinguish between artificial and machine-generated music.
\\[0pt]
\indent
As one of the most important core elements in music, mode plays an important role in judging music. Some relevant literature exists that examines the identification algorithm of Chinese modes, but sufficient research on identifying western modes remains to be seen; the context of judgement technology for analysing machine-generated artificial music through western modes is a particularly sparse area of research.
\\[0pt]
\indent
In our previous study, a mode-identification algorithm was designed \cite{1}, which can classify Chinese traditional modes by constructing a decision-making tree and judging the emotion in Chinese traditional music through identifying modes. The algorithm is consequently shown to have a fairly high accuracy rate for identifying traditional Chinese modes, and thus distinguishing whether or not it is indeed a traditional Chinese mode. While the algorithm’s judgement on traditional Chinese modes is fairly accurate, it also exhibits effective anti-interference performance and can successfully identify non-traditional Chinese modes. On this basis, some scholars have constructed a traditional music mode pattern based on traditional Chinese music theory \cite{3}, matching the traditional Chinese music modes. The findings indicate that the algorithm has quite a high accuracy rate in identifying traditional Chinese music modes and can distinguish between pentatonic and heptatonic modes.
\\[0pt]
\indent
In previous studies, we have proposed CFCS \cite{2}, the chord theory constructor based on the chord’s construction law and processing logic, and have designed a dynamic programming algorithm for the automatic composition of chords; this enables the realisation of mechanised automatic chord composition. Through experimentation in various cases, the algorithm has been proven to be feasible and effective.\\[0pt]
\indent
The article proposes OSC (Occidental Scale Constructor) based on a combination of research on traditional Chinese modes and CFCS chord composition function. By constructing the function to conduct mode analysis on monody, the article will make judgements on machine-generated and artificial music based on model stability and abnormal mode changes. Due to the subjectivity and territoriality of music, the range of the study will be limited to popular music based on natural major and minor tunes. The processing of modifier notes such as passing notes, neighbouring nodes, and nonessential notes will not be included.

\section{Approach}

The main technical issue that the article aims to resolve is the design of an algorithm that can distinguish between artificial and machine-generated music. The adopted technical proposal is to analyse melodic data through a set of western mode construction functions and subsequently make the distinction based on the analytical result.
\\[0pt]
\indent
As shown in Fig. \ref{fig:1},  the overall technical pattern of the research can be divided into three parts. Firstly, decode the MIDI byte through a MIDI preprocessing module and divide some characteristic series according to specific music rules. Secondly, analyse the preprocessed files through mode analysis mode to ascertain whether the melody adheres to basic music rules. Finally, identify the data in the last module in accordance with a man-machine identification module, to assess the probability of the melody being either man- or machine-made.

\subsection{MIDI File Preprocessing}
MIDI (Musical Instrument Digital Interface) was introduced in the 1980s to amend communication issues between electroacoustic musical instruments, and is currently the most widely accepted music standard format in the composition world; almost all modern music is created and composed using MIDI. As MIDI files usually contain a large amount of information, it is essential to preprocess the MIDI data used in our experiments. Preprocessing mainly involves extracting the scale based on the pitch of the MIDI file, thereby eliminating different interference notes by enumerating the filtration of characteristic intervals and statistical frequency to improve the accuracy of the final result.
\\[0pt]
\indent
As shown in Fig. \ref{fig:2}, the model mainly uses the music’s abstract information extracted from the MIDI files for subsequent calculation. It identifies the tracks in the MIDI (accompaniment, drumbeat, melody, polyphony, etc.) based on the established rules before classifying the music construction.
\\[0pt]
\indent
After the MIDI is decoded, the model obtains a series of abstract MIDI information. As demonstrated in Fig.\ref{fig:3},  where ‘0’, ‘1’, ‘n’ denotes the order of bars, MIDI information is then divided into bars and categorised after data cleansing through a series of classification layers. Finally, each MIDI track goes through modal orientation extraction.

\subsection{Modal Extractor}

The modal extractor extracts the possible mode set of each bar of preprocessed MIDI data through pre-established rules, making bar mode selections regarding the overall most orientated mode. The most frequently used method for extracting the tendentious set is to match the model exclusion mask based on the model rule library generated by OSC and deduce the possible model backward via calculation of the exclusive ones. (Fig.\ref{fig:4})

\subsection{Modal Rule Library and OSC}

Model is a form of organisation structure of music tones with a long-established history of use in practical music. When describing the concept of model, people typically take the pivot note of a model, i.e., the keynote, as the starting and finishing points. Other notes will be arranged in the form of a scale, based on the sequence of the pitch. This is known as modes.
\\[0pt]
\indent
The natural major and minor are the most common modes in the western modal system and in pop music to this day. The article proposes Occidental Scale Constructor (OSC) and constructs the model rule library based on the composition system of natural major and minor modes.

\begin{figure*}[htb]
	\centering
	\includegraphics[width=0.7\textwidth]{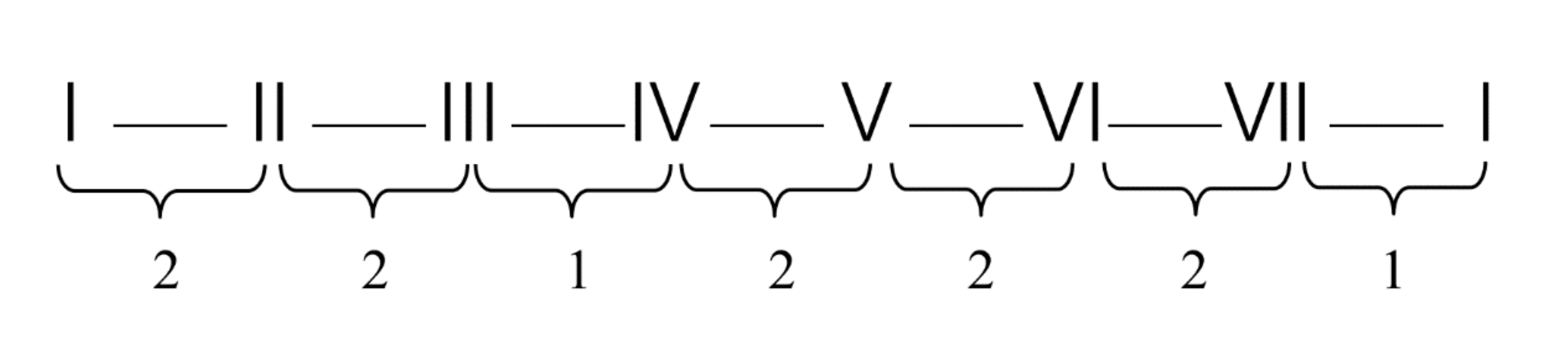}
	\caption{The composition principle of natural major}
	\label{fig:5}
\end{figure*}

\begin{figure*}[htb]
	\centering
	\includegraphics[width=0.7\textwidth]{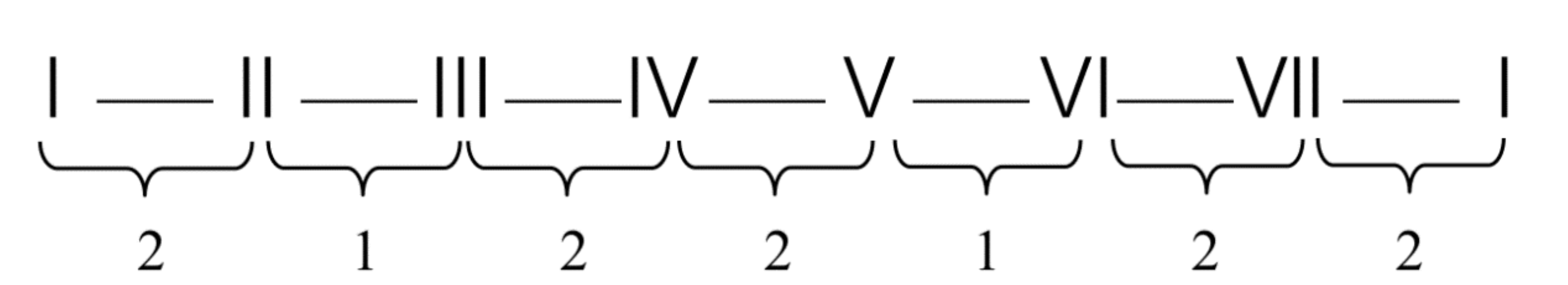}
	\caption{The composition principle of natural minor}
	\label{fig:6}
\end{figure*}

\subsubsection{The Constructor of Natural Major}
\indent

The natural major is a scale system consisting of two whole tones, a semitone, three whole tones, and a semitone. See Fig.\ref{fig:5}, where ‘2’ denotes a whole tone and ‘1’ denotes a semitone. Starting from any note, any scale system that is constructed in accordance with the aforementioned rules can be called a natural major system.

Based on the rules above, the construction function of the natural major can be formulated as:
\begin{equation}
\begin{aligned}
F_{Major}(S,O)=[&S+(O*12),S+2+(O*12),\\&S+4+(O*12),
S+5+(O*12),\\&S+7+(O*12),S+9+(O*12),\\&S+11+(O*12)]
\label{eq:1}
\end{aligned}
\end{equation}

Under the mapping relation F (function), S (step) in the function refers to any given sound level, while O (octave) represents the octave group. The natural major scale of current group can thus be constructed.

\subsubsection{The Constructor of Natural Minor}
\indent

The constitution system of the natural minor is a whole tone, a semitone, two whole tones, a semitone, and two whole tones. See Fig.\ref{fig:6}.

According to Eq. (\ref{eq:1}), the construction function of the natural minor key will therefore be:

\begin{equation}
\begin{aligned}
F_{Minor}(S,O)=[ &S+(O*12),S+2+(O*12),\\&S+3+(O*12), S+5+(O*12), \\
&S+7+(O*12),S+8+(O*12), \\
&S+10+(O*12)]
\label{eq:2}
\end{aligned}
\end{equation}

\begin{figure*}[htb]
	\centering
	\includegraphics[width=0.7\textwidth]{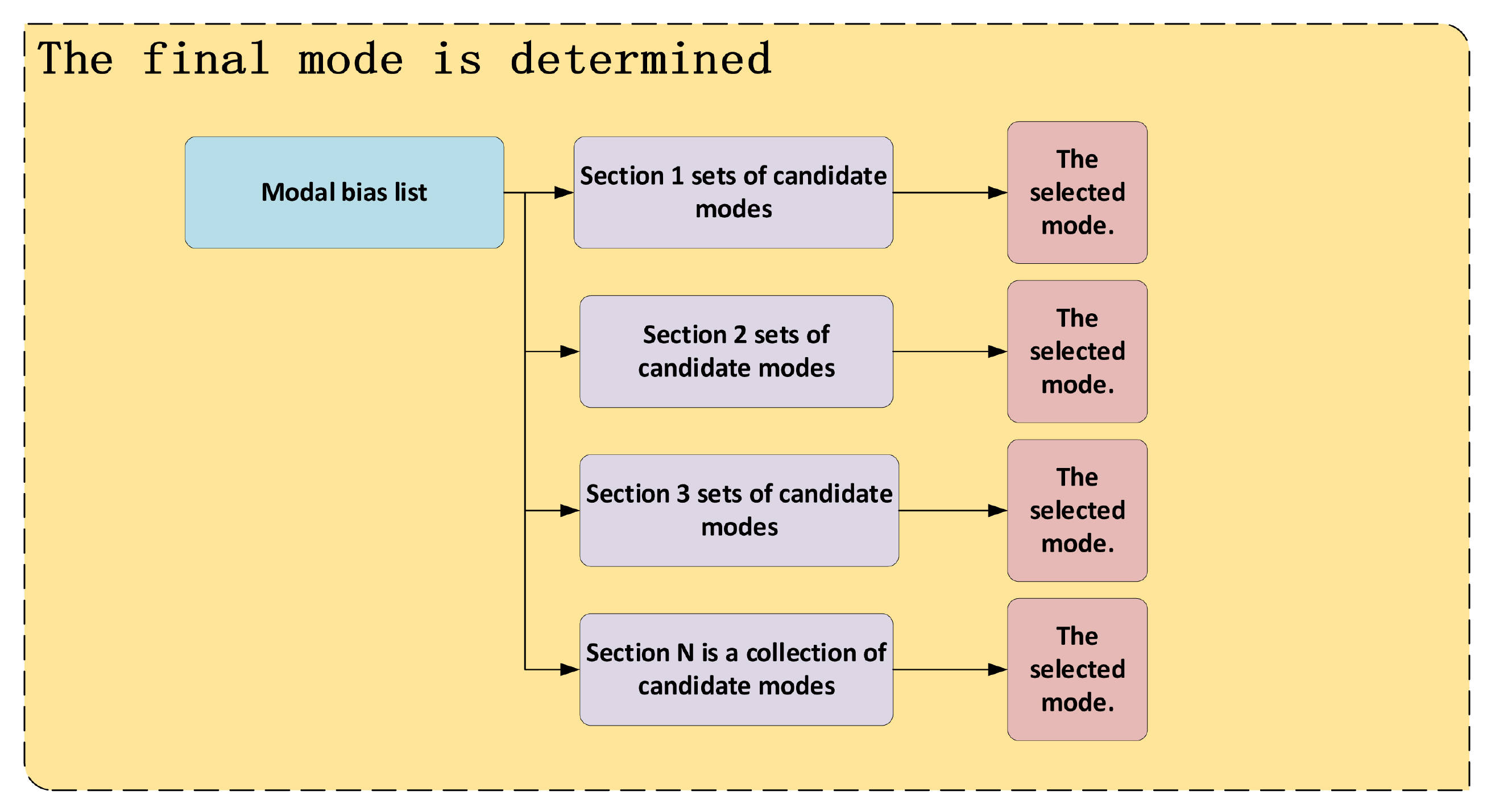}
	\caption{Mode determination}
	\label{fig:7}
\end{figure*}
\begin{figure*}[htb]
	\centering
	\includegraphics[width=0.7\textwidth]{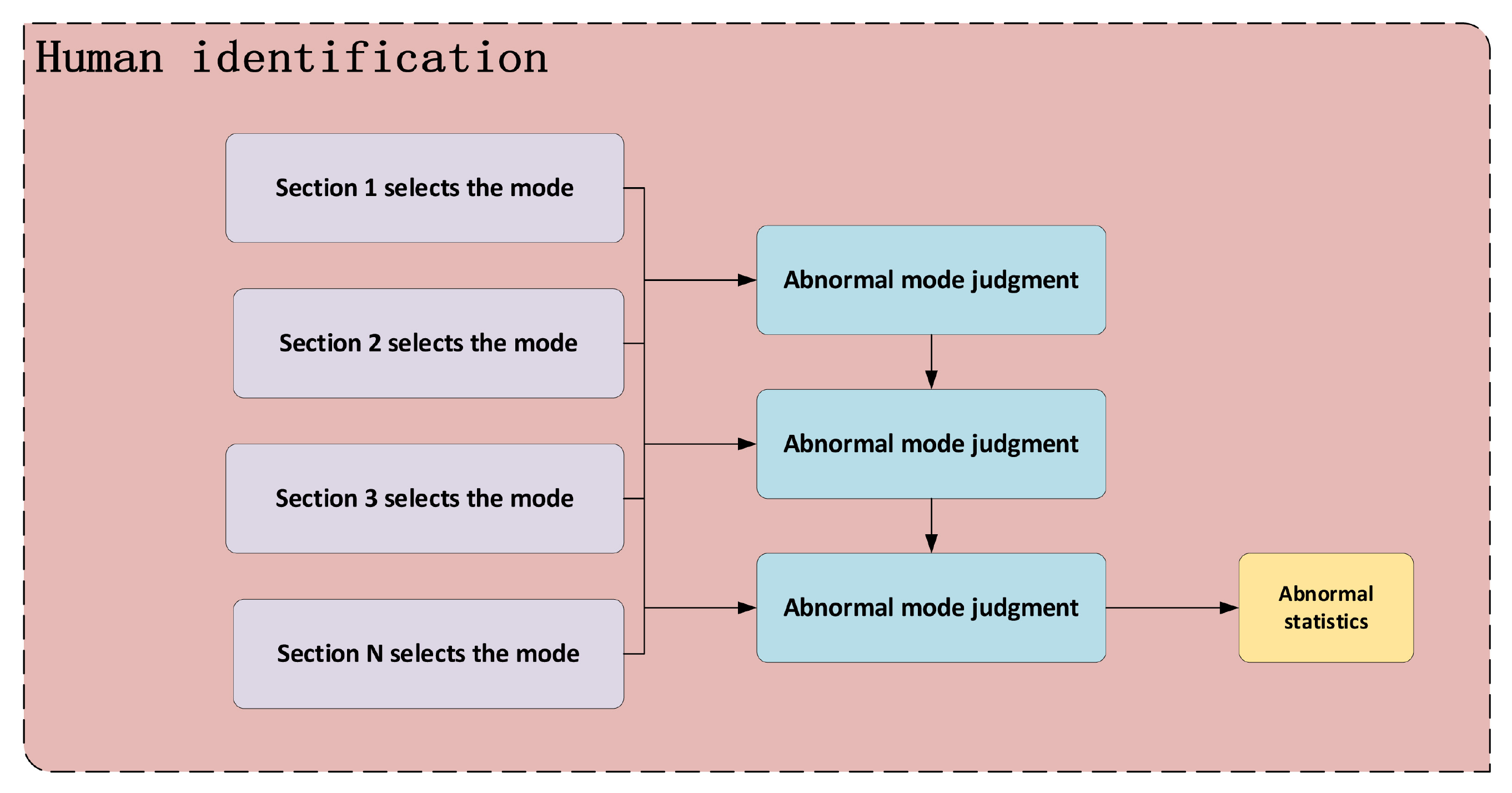}
	\caption{AI composition recognition algorithm}
	\label{fig:8}
\end{figure*}

\subsection{Mask Remove Algorithm}
It is extremely unlikely that the melody of a single bar would exhibit the complete scale of all models. For example, when there is any black key note, only C natural major can be excluded while all remaining models can still potentially become the dominative model of the entire piece. Based on this issue, it is possible to construct an excluding M (masking) for the melody of a given bar based on the constitution system of natural major and minor. Conducting model-exclusive calculations on the pitch is also an option, for the purpose of obtaining all variant models of the current bar before conducting a systematic analysis on all variant models and ascertaining the dominative model of the entire piece.

The mask sequence based on the major will be constructed as such:

\begin{equation}
\begin{aligned}
M_{major}(S,O)= [&S +1+(O*12), S +3+(O*12), \\
&S +6+(O*12), S +8+(O*12), \\
& S +10+(O*12)]
\label{eq:3}
\end{aligned}
\end{equation}

Compared with the natural major, the scale of natural minor elevates the fifth scale on the foundation of the natural major. Consequently, the mask sequence construction function of minor will be:

\begin{equation}
\begin{aligned}
M_{minor}(S,O)= [&S +1+(O*12), S +3+(O*12), \\
&S +6+(O*12), S +7+(O*12),  \\
&S +10+(O*12)]
\label{eq:4}
\end{aligned}
\end{equation}

If the scale in MMinor (S,O) is not evident in some bars, the affiliated minor of the major whose key note is S can be adopted as the alternative model of the current bar.

Under the mapping relation of the M (Mask), with given S (Step) and O (Octave), the exclusive sequence of the natural major that uses S as keynote can be obtained. When the pitch of the bar is in this sequence, we can exclude this model. Taking C natural major as an example, when the model is C natural major and S=0, then:

\begin{equation}\begin{aligned}
M_{major}=[&1+(O*12), 3+(O*12), 6+(O*12), \\
& 8+(O*12), 10+(O*12)]
\label{eq:5}\end{aligned}
\end{equation}

If in some bars, Pitch-13, it can calculate the scale of O based on the twelve-tone equal temperament. And when O=1, then:

\begin{equation}
M_{major}(0,1) = [13,15,18,20,22]
\label{eq:6}
\end{equation}

Therefore,

\begin{equation}
Pitch \in M_{major}(0,1)
\label{eq:7}
\end{equation}

According to the above results, it can be concluded that the current bar does not belong to C natural major. After excluding all impossible models based on each bar of the piece in its entirety, the set of all possible models of the current bar can be obtained. After statistically analysing all alternative models, the model’s tendency sequence can then be calculated. Based on the model tendency, it would be possible to filtrate the alternative models of all bars.

For example, through calculation, it is possible to determine that the model tendency sequence list of a piece is [C major, G major, D major…] and the alternative model set of the first bar is [G major, A major, E major…]. Consequently, if one were to make the choice based on the sequence in the list, the result would be G major. Likewise, by selecting the model for all bars, the model tendency of the whole piece would thus be obtained.(Fig.\ref{fig:7})

\begin{figure*}[htb]
	\centering
	\includegraphics[width=0.7\textwidth]{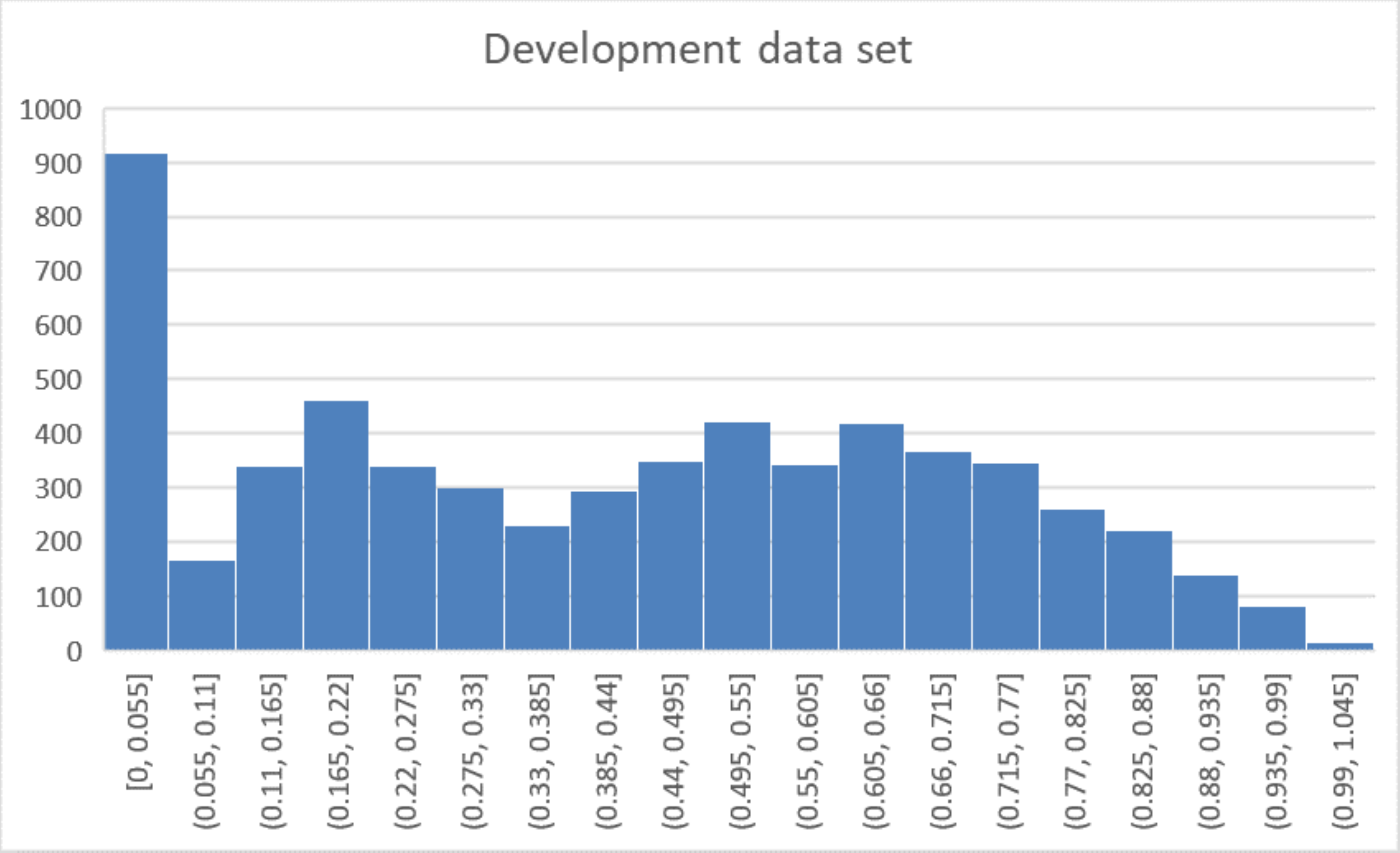}
	\caption{score distributtion of the development data}
	\label{fig:9}
\end{figure*}

\begin{figure*}[htb]
	\centering
	\includegraphics[width=0.7\textwidth]{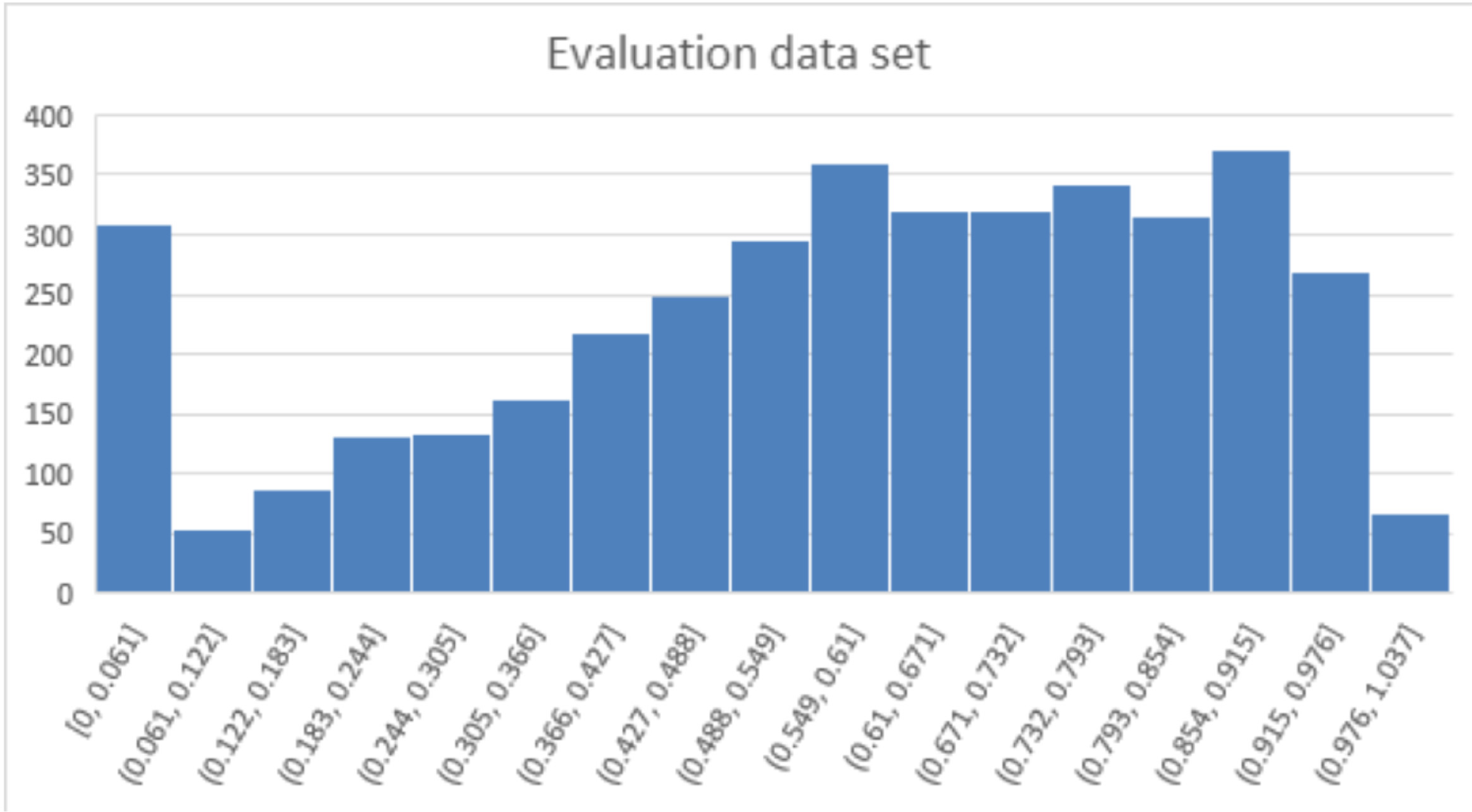}
	\caption{score distributtion of the evaluation data}
	\label{fig:10}
\end{figure*}

\subsection{AI composition recognition}
One of the most significant features of music is model stability. Although many musicians commit themselves to breaking the regular model system and discovering new creation techniques, mainstream music currently still adopts the stable model. Even the modulation or detune obeys certain rules and frequency. For example, modulation usually occurs between closely related models, as frequent or distant modulation would influence the stability of the music. Therefore, the article designs an algorithm to judge abnormal models and consequently attain the statistics of the abnormal model change, so as to judge the probability of the music being artificial or machine-made.
\\ [0pt]\indent
Fig.\ref{fig:8} illustrates the technological flow chart that can be adopted to judge man-made or machine-made property through abnormal model change. This abnormal model change usually takes the form of unconventional modulation or with uncertain model. For instance, the models of the bars in one melody are identified as [C, C, C, G, G, E, A, B, F, A, C]. The former five bars are [C, C, C, G, G]. The transmission from C to G belongs to close modulation, so there is no abnormal model change. However, the models [E, A, B, F, A, C] that follow it are not closely related; this case can therefore be judged as abnormal model change. Six instances of abnormal model change can be identified in this melody, while there are ten instances when the model can be modified. Thus, the output score of the melody is 6/10=0.6.

\section{Experiment}

The data used are provided by CSMT. The development dataset contains 6000 MIDI files with monophonic melodies generated by artificial intelligence algorithms. The tempo is between the 68bpm and 118bpm (beat per minute). The length of each melody is 8 bars, and the melody does not necessarily include complete phrase structures. The evaluation dataset contains 4000 MIDI files with exact configurations of development dataset with two exceptions: 1) A number of melodies composed by human composers are added, 2) There are a number of melodies generated by algorithms with minor difference compared to the algorithms in the development dataset.
\\ [0pt]\indent
Experimental results on CSMT datasets indicate that the score distributtion of the development data is obviously at a low level(Fig.\ref{fig:9}), while the score distributtion of the evaluation data is obviously at a high level(Fig.\ref{fig:10}).
\\ [0pt]\indent
Through experiments on 10,000 samples, our algorithm shows a successful identification performance on the judgment of man- or machine-made works. However, complex composing techniques and the evaluation of the time value of notes are not be included. Under the circumstance of short duration time, the melody created by human and machine can not be clearly judged by composition techniques and abstract rules such as musical form structure. A small number of melody pieces can not be clearly judged even by professionals. However, considering that the melody itself has a certain flexibility, there is no strict unified standard, so the experimental results prove that the algorithm is effective and feasible.

\section{Conclusion}

Starting from the music mode recognition and the essence of the music, the article proposes Occidental Scale Constructor based on the CFCS chord constructor. The article  also constructs a mode-based music-rule-identifying algorithm through combining OSC with the mask remove algorithm, which will identify the mode stability and abnormal mode change, to judge whether the piece is machine-generated. Experimental results on CSMT datasets demonstrate the algorithm to have a successful identification ability of machine-generated music. The algorithm will also provide some technological reference to the benign development of the music copyright and artificial intelligent music.

{\small
\bibliographystyle{ieee_fullname}
	\bibliography{mybib}
}

\end{document}